\documentclass[letterpaper, 10 pt, conference]{ieeeconf}
\IEEEoverridecommandlockouts
\usepackage{times}

\usepackage{multicol}
\usepackage[bookmarks=true]{hyperref}
\usepackage{graphicx}

\makeatletter
\let\NAT@parse\undefined
\makeatother
\usepackage[numbers,sort,compress]{natbib}

\usepackage{multirow}
\usepackage{multicol}
\usepackage{graphicx}
\usepackage{xspace}
\usepackage{xcolor}
\usepackage{caption}
\usepackage{wrapfig}
\usepackage{bbding}  
\usepackage{pifont}  

\usepackage{booktabs}
\usepackage{float}
\usepackage{amsmath}  
\usepackage{amssymb}  
\usepackage{bm}
\usepackage{dsfont}
\newcommand{\ours}[0]{AdaMimic\xspace}

\usepackage{hyperref}
\usepackage[capitalise, nameinlink]{cleveref}

\newcommand{\paragraphbegin}[1]{\vspace{0.0in}\noindent\textbf{#1}}

\usepackage{colortbl}
\definecolor{ourcolor}{HTML}{99e0eb}
\definecolor{ourblue}{HTML}{27a2c3}

\definecolor{tablecolor}{HTML}{ccf2f5} 

\definecolor{tablecolor2}{HTML}{ffcdb4}
\definecolor{citecolor}{HTML}{fe7b5b}
\definecolor{grey}{rgb}{0.9, 0.9, 0.9}
\usepackage{amssymb}

\usepackage{listings}
\definecolor{gred}{rgb}{0.859,0.267,0.216}
\definecolor{ggreen}{rgb}{0.059,0.616,0.345}

\definecolor{deepblue}{HTML}{27a2c3}

\definecolor{deepred}{HTML}{fe7b5b}

\newcommand{\eg}[0]{\textit{e.g.},\xspace}

\usepackage[font=small,labelfont=bf]{caption}

\usepackage[font=footnotesize,labelfont=bf]{caption}

\definecolor{citecolor}{HTML}{faa700} 
\definecolor{lblue}{HTML}{ffb114} 
\definecolor{ogreen}{HTML}{2E7D32}
\definecolor{bred}{HTML}{BF360C}
\definecolor{newbrown}{HTML}{795548}

\definecolor{citecolor}{HTML}{c03d3e}
\hypersetup{
    colorlinks=true,
    linkcolor=red,
    filecolor=blue,      
    urlcolor=blue,
    citecolor=blue,
}

\newcommand{\ourrow}{\rowcolor{gray!7}}

\newcommand{\ci}[1]{\tiny{\textcolor{gray}{$\pm #1$}}}

\usepackage{multicol}
\usepackage{multirow}
\usepackage{colortbl}
\usepackage{booktabs}   
\usepackage{bbding} 
\usepackage{graphicx}
\let\labelindent\relax 
\usepackage{enumitem}

\usepackage{bbding}

\usepackage{pifont}
\usepackage{xfrac}
\usepackage{tikz}

\usepackage[para,online,flushleft]{threeparttable}
\newcommand{\thickcdot}{
  \mathbin{\raisebox{0.15ex}{\scalebox{0.8}{$\bullet$}}}%
}

\usepackage{amsfonts}
\usepackage{bm}
\usepackage{amsmath}
\usepackage{mathtools}
\usepackage{multirow}
\usepackage{booktabs}
\usepackage{caption}
\usepackage{wrapfig,lipsum,booktabs}
\usepackage{caption}
\usepackage{bbm}
\usepackage{multirow}
\usepackage{pifont}
\usepackage{tablefootnote}
\usepackage{threeparttable} 

\newcommand{\metlocal}{E_{\mathrm{l-bpe}}^\mathrm{dense}}
\newcommand{\metglobal}{E_\mathrm{g-bpe}^\mathrm{sparse}}
\newcommand{\metsmooth}{E_\mathrm{smth}^\mathrm{dense}}
\newcommand{\metsuccess}{\mathrm{Success}}

\newcommand{\phaseadapter}{\pi_{\mathrm{phase}}^\Delta}
\newcommand{\deltaphase}{\Delta \phi_k^{\Delta}}
\newcommand{\adaptivephase}{\Delta \phi_k^{\mathrm{ada}}}

\newcommand{\trackingadapter}{\pi_{\mathrm{track}}^\Delta}
\newcommand{\deltatrack}{\bs{a}_k^{\Delta}}
\newcommand{\adaptivetrack}{\bs{a}_k^{\mathrm{ada}}}

\newcommand{\bs}[1]{\boldsymbol{#1}}

\definecolor{goldenyellow}{rgb}{0.99, 0.76, 0.0}
\usepackage{cleveref}
\usepackage{float}

\begin{document}

\title{\LARGE \bf Towards Adaptable Humanoid Control via Adaptive Motion Tracking}


\author{
\authorblockN{
Tao Huang$^{2,1}$ \quad
Huayi Wang$^{1,2}$ \quad
Junli Ren$^{1}$ \quad
Kangning Yin$^{1,2}$ \quad
Zirui Wang$^{1}$ \quad
Xiao Chen$^{1}$ \quad \\
Feiyu Jia$^{1}$ \quad
Wentao Zhang$^{1}$ \quad
Junfeng Long$^{1}$ \quad
Jingbo Wang$^{1,\dag}$ \quad
Jiangmiao Pang$^{1,\dag}$
}
\authorblockA{
\textsuperscript{1}Shanghai AI Laboratory \quad \textsuperscript{2}Shanghai Jiao Tong University  \quad \textsuperscript{$\dagger$}Equal Advising\\
Website: \href{https://taohuang13.github.io/adamimic.github.io/}{\texttt{adamimic.github.io}} \quad Code: \href{https://github.com/InternRobotics/AdaMimic}{\texttt{https://github.com/InternRobotics/AdaMimic}}
}
}

\twocolumn[{%
\renewcommand\twocolumn[1][]{#1}%
\maketitle
\vspace{-0.6cm}
\begin{center}
    \centering
    \captionsetup{type=figure}
     \includegraphics[width=1\textwidth]{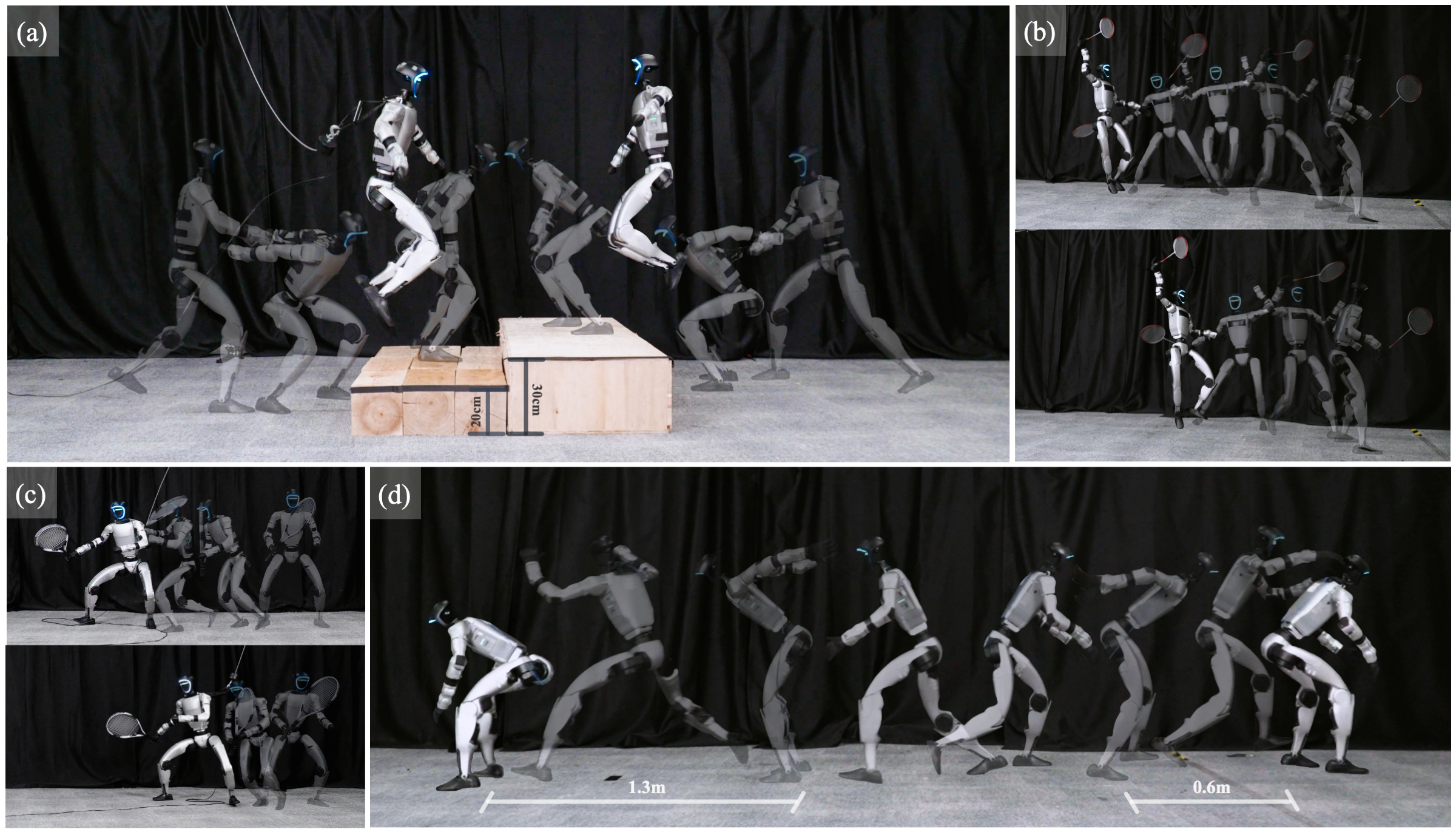}
     \vspace{-0.17in}
    \caption{\textbf{Overview.} Our method, \textbf{\ours} (adaptive motion tracking), achieves agile humanoid whole-body adaptation from only a single reference motion while consistently preserving the underlying motion patterns. This enables adaptable control across diverse tasks, as illustrated by the varied outcomes: (a) higher jumping height, (b) extended movement in badminton hitting, (c) extended movement in tennis hitting, and (d) longer jumping distance.}
    \label{fig:teaser}
\end{center}
\vspace{0.12in}
}]


\begin{abstract}
Humanoid robots are envisioned to adapt demonstrated motions to diverse real-world conditions while accurately preserving motion patterns. Existing motion prior approaches enable well adaptability with a few motions but often sacrifice imitation accuracy, whereas motion-tracking methods achieve accurate imitation yet require many training motions and a test-time target motion to adapt. To combine their strengths, we introduce \ours, a novel motion tracking algorithm that enables adaptable humanoid control from a single reference motion. To reduce data dependence while ensuring adaptability, our method first creates an augmented dataset by sparsifying the single reference motion into keyframes and applying light editing with minimal physical assumptions. A policy is then initialized by tracking these sparse keyframes to generate dense intermediate motions, and adapters are subsequently trained to adjust tracking speed and refine low-level actions based on the adjustment, enabling flexible time warping that further improves imitation accuracy and adaptability. We validate these significant improvements in our approach in both simulation and the real-world Unitree G1 humanoid robot in multiple tasks across a wide range of adaptation conditions (\cref{fig:teaser}). Videos and code are available on our \href{https://taohuang13.github.io/adamimic.github.io/}{project page}.
\end{abstract}

\IEEEpeerreviewmaketitle

\section{Introduction}
Humans are good at acquiring whole-body skills by first mimicking experts and then adapting to new situations. For example, a tennis player reproduces the expert stroke pattern given different ball positions. Humanoid robots are expected to have a similar capability: adapt from reference motions to diverse real-world conditions, while accurately imitating motion patterns. We refer to this ability as adaptable humanoid control from reference motions.

Approaches to this motion-based adaptable control fall into two main paradigms. Methods that incorporate reference motions as prior into reinforcement learning (RL) enable adaptation beyond the data~\cite{peng2021amp,Ma2025StyleLocoGA,Liao2025BeyondMimicFM,Xue2025LeVERBHW,Allshire2025VisualIE}, but they often sacrifice imitation accuracy or require extensive reward tuning. Motion tracking methods, in contrast, can reproduce reference motions accurately with lower reward engineering burden~\cite{peng2018deepmimic,he2025asap,Xie2025KungfuBotPH,Zhang2025HuBLE,He2024OmniH2OUA,He2024HOVERVN,Shi2025AdversarialLA,Xue2025AUA,Cheng2024ExpressiveWC,Zhuang2025EmbraceCH,Li2025CLONECW}. However, their adaptability may be limited by the dependence on large-scale training motions to cover diverse conditions and on possibly unavailable reference motions for each test-time deployment~\cite{Chen2025GMTGM,Yin2025UniTrackerLU}. How to combine the strengths of both paradigms—accurate imitation and broad adaptability—remains an open challenge.

In this work, we take an initial step towards this challenge by introducing \ours,  novel motion tracking algorithm that enables humanoid robots to adapt from a single reference motion while accurately preserving motion patterns. To reduce data dependence while supporting adaptation, we first generate an augmented dataset by sparsifying the reference motion into keyframes and editing a few keyframes with minimal physical assumptions, preserving the essential local pattern of the reference data. In the first stage, the policy is trained to track these sparse keyframes, producing dense intermediate motions that maintain local patterns to the reference. In a second stage, two adapters are jointly learned: a phase adapter that modulates motion speed, and a tracking adapter that compensates for the low-level actions based on the adjustment. This two-stage design enables flexible time warping~\cite{witkin1995motion,hsu2007guided}, enhancing both imitation accuracy and adaptability. The resulting policies can be deployed on hardware without requiring additional reference motions.

Extensive evaluations across diverse tasks and conditions, both in simulation and on the real-world Unitree G1 humanoid robot, demonstrate that \ours outperforms existing methods. We overview of its real-world performance in~\cref{fig:teaser} and summarize our core contributions as follows:
\begin{itemize}[leftmargin=4mm]
    \item We introduce \ours, a novel motion tracking algorithm that enables humanoid robots to adapt from a single reference motion while accurately preserving key patterns.
\item We validate the effectiveness of \ours in extensive simulations, demonstrating significantly improved imitation accuracy and adaptability than existing methods.

\item We deploy the trained policies on the real-world Unitree G1 humanoid robot, showing good performance across wide adaptation conditions in multiple tasks.
\end{itemize}

\begin{figure*}[tbp]
    \centering
    \includegraphics[width=\linewidth]{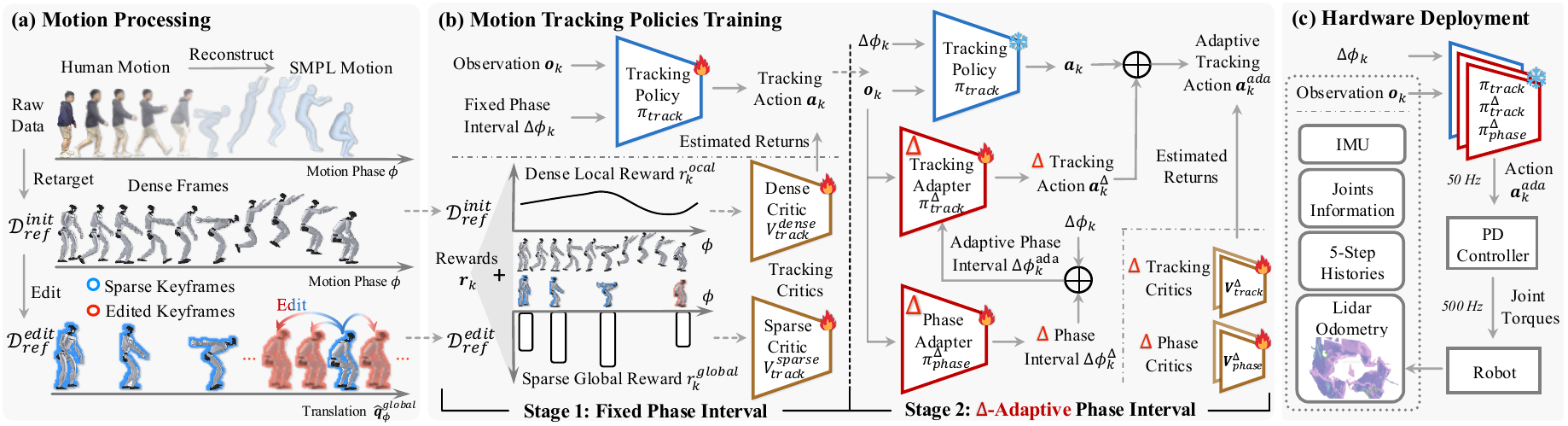}
    \caption{\textbf{Method overview.} (a) Human motions are reconstructed into SMPL motions via GVHMR~\cite{Shen2024WorldGroundedHM} and retargeted to the humanoid robot. Sparse keyframes are then selected and edited to form an augmented dataset for adaptive tracking. (b) Based on this dataset, \ours\ first trains a tracking policy with fixed phase intervals and double critics for sparse global tracking and dense local tracking rewards, followed by phase and tracking adapters that enable effective time warping for improved tracking performance. (c) The resulting policies can be directly deployed on the real Unitree G1 robot.}
    \label{fig:method}
    \vspace{-0.1in}
\end{figure*}
\section{Related Work}
\subsection{Adaptable Humanoid Control}
Adaptation to diverse real-world conditions is crucial in humanoid control, spanning both locomotion~\cite{radosavovic2024real,li2024reinforcement,zhuang2024humanoid,long2024learning,Li2025HoldMB,wang2025beamdojo,He2025AttentionBasedME,Gu2024AdvancingHL} and whole-body manipulation~\cite{Fu2024HumanPlusHS,Lin2025SimtoRealRL,Lu2024MobileTeleVisionPM,Li2025AMOAM,Zhang2025FALCONLF,Ben2025HOMIEHL,Ze2025TWISTTW,liu2024opt2skill}. Approaches to these tasks often rely on sophisticated reward engineering and carefully designed sim-to-real transfer pipelines. In parallel, another line of work leverages human motion references to acquire adaptable whole-body skills~\cite{peng2021amp,Ma2025StyleLocoGA,Liao2025BeyondMimicFM,Xue2025LeVERBHW,Allshire2025VisualIE,Zhuang2025EmbraceCH}. Their methods typically utilize motion data as priors, prioritizing adaptability over the strict preservation of original motion patterns. In contrast, our work emphasizes accurate tracking alongside adaptation.  

\subsection{Humanoid Motion Tracking}
Motion tracking is proven effective for accurate imitation of a single reference motion~\cite{peng2018deepmimic,he2025asap,Xie2025KungfuBotPH,Zhang2025HuBLE}. However, its adaptability is often constrained by the scale of data. Recent advances address this limitation by training from massive public datasets~\cite{He2024OmniH2OUA,He2024HOVERVN,Chen2025GMTGM,Shi2025AdversarialLA,Xue2025AUA,Yin2025UniTrackerLU,Cheng2024ExpressiveWC,Ji2024ExBody2AE,Zhuang2025EmbraceCH,Li2025CLONECW}. While such approaches demonstrate impressive generalization, they typically require substantial data curation, rely on teleoperation or pre-collected reference motions during deployment, and may face challenges in maintaining tracking accuracy for highly agile behaviors. In contrast, our work explores how adaptive tracking can be achieved in agile tasks from only a single reference motion, without the need for additional pre-collected trajectories.

\subsection{Motion Adaptation}
A classical approach to adapting humanoid motions is motion editing, which can be achieved either through spacetime constraints~\cite{gleicher1997spacetime,witkin1995motion,Lee1999AHA,popovic1999physically} or data-driven generation~\cite{Harvey2020RobustMI,Delmas2023PoseFixC3,Qin2022MotionIV}. While both paradigms enable reusing a single motion as input, they often face limitations in physical plausibility caused by over-assumed constraints~\cite{popovic1999physically,liu2005learning} or under-assumed dynamics~\cite{Yuan2022PhysDiffPH}, limiting real-world deployment. This has motivated physics-based approaches that integrate simulation and reinforcement learning to improve motion plausibility~\cite{zargarbashi2024robotkeyframing,lee2021learning,Huang2024DiffuseLocoRL,Huang2025DiffuseCLoCGD}. However, such methods often depend on large-scale datasets for interpolation or rule-based plausibility heuristics. Inspired by these two lines of work, we explore an alternative strategy: selecting and editing keyframes from a single motion, and then leveraging RL-based tracking to generate physically plausible in-between motions.

\section{Background: Humanoid Motion Tracking}
We formulate humanoid motion tracking as a goal-conditioned reinforcement learning (RL) problem within the framework of a Markov decision process (MDP;~\cite{puterman2014markov}). The objective is to learn a tracking policy $\pi_{\mathrm{track}}$ that accurately reproduces a reference motion $\hat{\bs{q}}$. 

\subsubsection{Motion representation} 
The retargeted reference motion $\hat{\bs{q}}$ is sampled from a reference dataset $\mathcal{D}_{\mathrm{ref}}^{\mathrm{init}}$, which initially contains a single motion~\cite{peng2018deepmimic,he2025asap}. Each motion trajectory consists of global poses $\hat{\bs{q}}^{\mathrm{global}}$ (\eg position and orientation of robot bodies) and local poses $\hat{\bs{q}}^{\mathrm{local}}$ (\eg joint angles). A normalized phase variable $\phi \in [0,1]$ parameterizes the whole reference motion:
$\hat{\bs{q}}_\phi = \big( \hat{\bs{q}}^{\mathrm{global}}_\phi, \hat{\bs{q}}^{\mathrm{local}}_\phi \big),$
which advances discretely at each timestep $k$ as:
\begin{equation}
    \phi_k = \phi_{k-1} + \Delta\phi_k,\quad \Delta\phi_k = \Delta t \;/\; T_{\hat{\bs{q}}},
\end{equation}
where $\Delta t$ is typically the simulation timestep and $T_{\hat{\bs{q}}}$ is the duration of the reference motion.

\subsubsection{Observations and actions}
At timestep $k$, the observation is defined as $
    \bs{o}_k = \big[\hat{\bs{q}}_{\phi_k},\bs{q}_{\phi_k}, \dot{\bs{\theta}}_k,  \dot{\bs{\omega}}_k, \phi_k\big],$
where $\bs{q}_{\phi_k}$ denotes current global and local robot poses, $\dot{\bs{\theta}}$ the joint velocities, and $\dot{\bs{\omega}}_k$ the base angular velocity.  
The policy produces a tracking action $\bs{a}_k \sim \pi_{\mathrm{track}}( \cdot| \bs{o}_k,\Delta\phi_k)$ conditioned on the current observation, the reference motion, and the fixed phase interval. The action is the target of a PD controller during a control period.

\subsubsection{Rewards and objectives} 
At timestep $k$, the rewards $\bs{r}_k = (r_k^{\mathrm{global}}, r_k^{\mathrm{local}})$ are aggregated from multiple terms that evaluate different aspects of tracking performance. For clarity, we group them into two categories: global-level rewards $r_k^{\mathrm{global}}$ and local-level rewards $r_k^{\mathrm{local}}$. The former encourages accurate tracking of global trajectories, while the latter focuses on matching local motion patterns. The tracking policy $\pi_{\mathrm{track}}$ is then optimized to maximize the expected return:
\begin{equation}\label{eq:objective-naive} 
    \max_{\pi_{\mathrm{track}}} \;\; 
    \mathbb{E}_{\pi_{\mathrm{track}}}\left[ \sum_{k=0} \gamma^k \, (\bs{w}\cdot \bs{r}_k) \;\middle|\; \Delta \phi_k, \hat{\bs{q}} \sim 
    \mathcal{D}_{\mathrm{ref}}^{\mathrm{init}} \right], 
\end{equation}
where $\bs{w}$ denotes the weighting between the two reward groups, and $\gamma\in[0,1)$ is the discount factor.

\section{Method: Adaptive Motion Tracking}\label{sec:method}
\subsection{Problem Reformulation}\label{subsec:formulation}
The key insight of adaptive motion tracking is to extend the standard motion tracking problem by allowing variations in global trajectories while strictly preserving the local motion pattern of a given motion. Formally, starting from an initial reference dataset $\mathcal{D}_{\mathrm{ref}}^{\mathrm{init}} = \{\hat{\bs{q}}\}$ that contains a single motion trajectory, we assume access to an augmented dataset $\mathcal{D}_{\mathrm{ref}}^{\mathrm{edit}} = \{\hat{\bs{q}}^{\mathrm{edit}}_i\}_{i=0}^{M}$. Each edited motion $\hat{\bs{q}}^{\mathrm{edit}}_i$ shares the same local joint trajectories as the original motion:
\begin{equation}
    \hat{\bs{q}}^{\mathrm{edit,\,local}}_{i,\phi_k} = \hat{\bs{q}}^{\mathrm{local}}_{\phi_k}, \quad \forall \, \hat{\bs{q}}^{\mathrm{edit}}_i \in \mathcal{D}_{\mathrm{ref}}^{\mathrm{edit}}, \;\; \forall \, \phi_k\in[0,1],
\end{equation}
while the global component $\hat{\bs{q}}^{\mathrm{edit,\,global}}_{\phi_k}$ may vary to reflect different displacements or base translations. For clarity, we interchangeably refer to $\hat{\bs{q}}^{\mathrm{edit}}_{i}$ and $\hat{\bs{q}}$ without causing ambiguity in the rest of the paper.

The tracking policy $\pi_{\mathrm{track}}$ is then optimized to reproduce motions sampled from $\mathcal{D}_{\mathrm{ref}}^{\mathrm{edit}}$:
\begin{equation}\label{eq:objective-stage1}
    \max_{\pi_{\mathrm{track}}} \;\; 
    \mathbb{E}_{\pi_{\mathrm{track}}}\left[ \sum_{k=0} \gamma^k \, (\bs{w}\cdot \bs{r}_k) \;\middle|\; \Delta \phi_k,\hat{\bs{q}} \sim \textcolor{blue}{\mathcal{D}_{\mathrm{ref}}^{\mathrm{edit}}} \right].
\end{equation}
This formulation abstracts away the specific mechanism for constructing $\mathcal{D}_{\mathrm{ref}}^{\mathrm{edit}}$, which will be discussed below. Adaptation is achieved by specifying a different reference motion. 

\begin{figure}[t]
    \centering
    \includegraphics[width=1\linewidth]{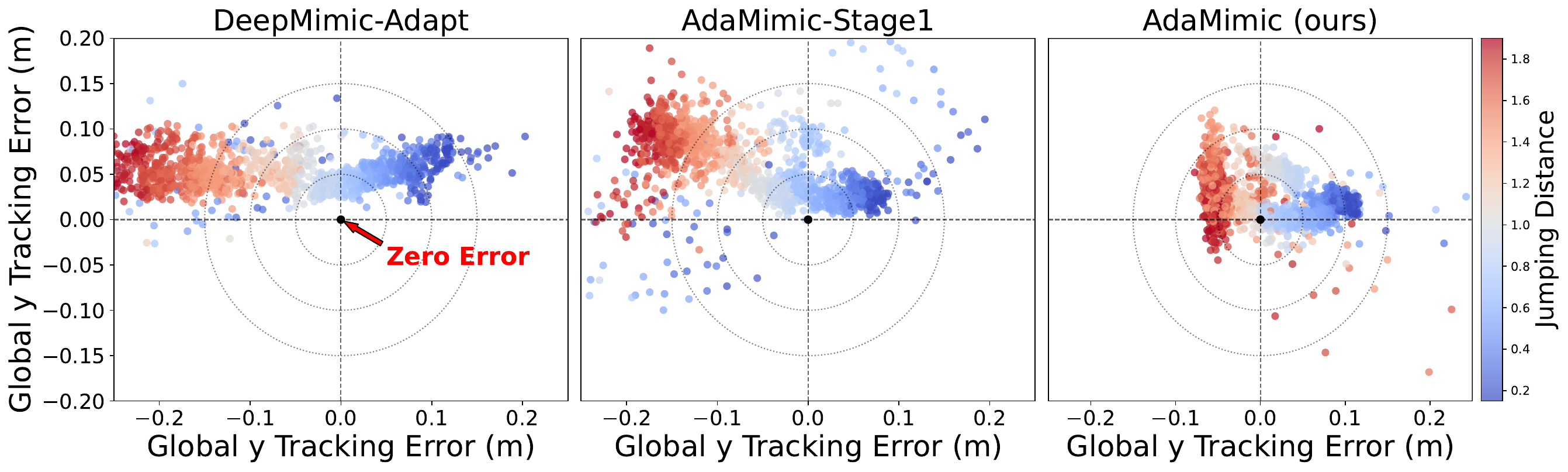}
    \includegraphics[width=1\linewidth]{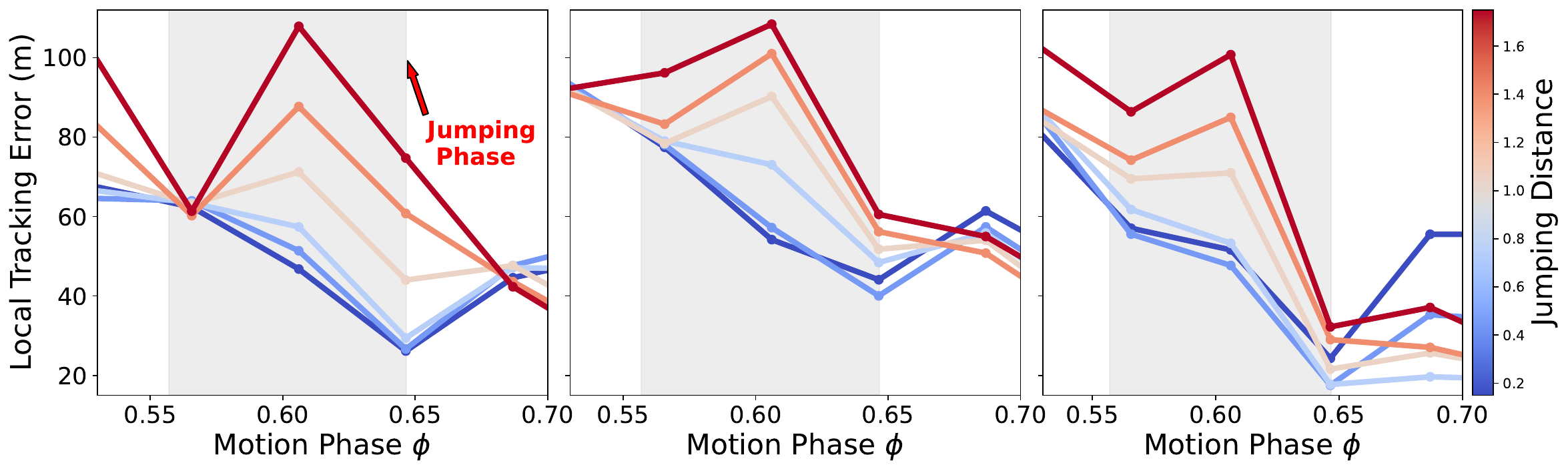}
    \caption{\textbf{Motivations of keyframing and adapters.} (Top) Global tracking errors at the landing moment of far jumping indicate that \ours\ outperforms baselines—DeepMimic-Adapt, which augments motions with rules, and AdaMimic-Stage1, which tracks at a fixed speed—by incorporating sparse keyframes and adaptive time warping to improve physical plausibility. (Bottom) Local tracking errors further verify the improvements.}
    \label{fig:stage2-motivation}
    \vspace{-0.1in}
\end{figure}

\subsection{Keyframing and Editing: $\mathcal{D}_{\mathrm{ref}}^{\mathrm{init}} \rightarrow \mathcal{D}_{\mathrm{ref}}^{\mathrm{edit}}$}\label{subsec:motion-editing}

A classical approach to construct $\mathcal{D}_{\mathrm{ref}}^{\mathrm{edit}}$ is to sparsely edit keyframes and interpolate the intermediate frames under trajectory-level consistency constraints~\cite{witkin1995motion,gleicher1997spacetime}, denoted as $\mathcal{D}_{\mathrm{ref}}^{\mathrm{rule}}$. However, such optimization-based methods may yield physically implausible motions, which may impede hardware deployment (see~\cref{fig:real_snapshot}), due to the absence of dynamics~\cite{popovic1999physically,liu2005learning}. Inspired by these works, we also adopt a keyframe-based editing scheme to preserve the global motion structure, but go beyond their limitations by using RL to generate more dynamically plausible intermediate frames, drawing from the ideas of ~\cite{zargarbashi2024robotkeyframing,lee2021learning,gleicher2001motion}.

\begin{figure*}[tbp]
    \centering
    \includegraphics[width=\linewidth]{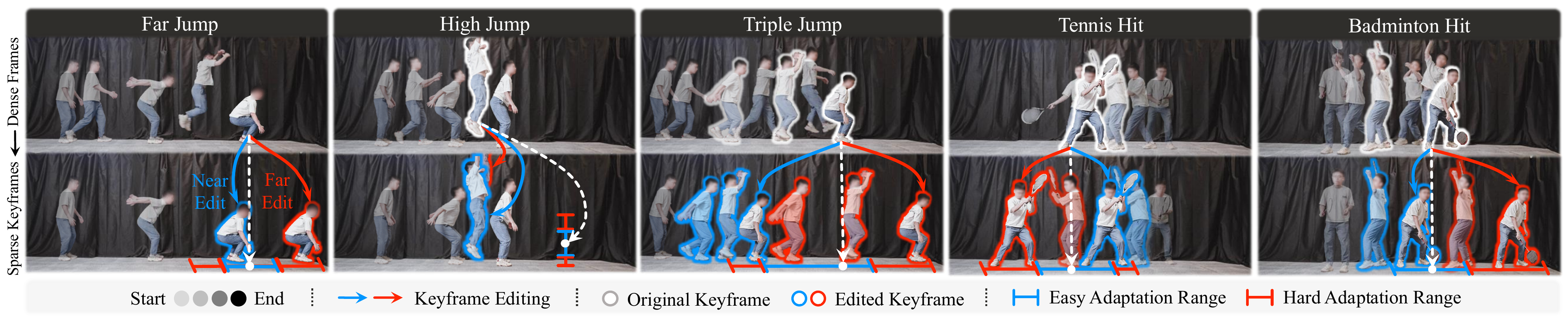}
    \caption{\textbf{(Top) Task visualization:} Five representative motions used as input for humanoid retargeting.
    \textbf{(Bottom) Keyframing and editing:} Sparse keyframes are extracted from each motion, and few selected ones are further edited to enable adaptation. Colors denote adaptation difficulty relative to the original keyframe (\textcolor{gray}{gray}): \textcolor{blue}{blue} indicates easy adaptation cases, while \textcolor{red}{red} indicates hard adaptation cases. The adaptation ranges are presented in~\cref{tab:adaptaion_range}.}
    \label{fig:difficulty-generalization}
    \vspace{-0.1in}
\end{figure*}

Concretely, we select $N$ keyframes, some of which are associated with semantic contexts (e.g., start, take-off, and landing in far jumping), and denote their phases as $\bs{\Phi}^{\mathrm{key}}=\{\phi^\mathrm{key}_0,\phi^\mathrm{key}_1,\dots,\phi^\mathrm{key}_{N-1}\}$. A subset $\bs{\Phi}^\mathrm{edit}\subset\bs{\Phi}^{\mathrm{key}}$ is then chosen for editing, with the principle of introducing as few physical assumptions as possible. For each edited keyframe at phase $\phi_k\in\bs{\Phi}^\mathrm{edit}$, we apply a transformation to its global pose based on a variable $\psi_i$ (e.g., jumping distance):
\begin{equation}
    \hat{\bs{q}}^{\mathrm{edit,\,global}}_{i,\phi_k} = f_\mathrm{edit}\big(\hat{\bs{q}}^{\mathrm{global}}_{\phi_k},\psi_i\big),
    \quad \forall \, \phi_k \in \bs{\Phi}^\mathrm{edit},
\end{equation}
while keeping the local joint path unchanged. This editing function $f_\mathrm{edit}$ is task-dependent; for instance, in far jumping, it may translate post-landing frames forward or backward. The resulting set of edited keyframes forms $\mathcal{D}_{\mathrm{ref}}^{\mathrm{edit}}$, which serves as reference motions for adaptive motion tracking. The whole editing process is illustrated in~\cref{fig:difficulty-generalization}.

\paragraphbegin{Remark:} The current scope of tasks we consider is limited to those where the keyframes $ \bs{\Phi}^{\mathrm{key}}$ and editing function $f_\mathrm{edit}$ can be defined in a relatively straightforward and task-specific way. Extending to more tasks requires developing more general mechanisms for constructing $ \bs{\Phi}^{\mathrm{key}}$ and $f_\mathrm{edit}$, which we leave as an important direction for future work.


 \subsection{Stage 1: Motion Tracking with Fixed Phase Interval}\label{subsec:training-stage1}
Given the edited motions, we first employ existing motion-tracking algorithms~\cite{he2025asap,peng2018deepmimic} to train a tracking policy $\pi_\mathrm{track}$ using a fixed phase interval $\Delta \phi$. This stage aims to provide the agent with an initial capability of imitating the edited keyframe motions under a simple and stable training setup.

\subsubsection{Sparse global reward}
To ensure global-space motion alignment, we design a sparse global reward function:
\begin{equation}
    r_k^{\mathrm{global}} = \mathcal{R}^\mathrm{global}_{\mathrm{track}}(\bs{q}_{\phi_{k+1}}^\mathrm{global},  \hat{\bs{q}}^{\mathrm{\,global}}_{\phi_{k+1}}) \cdot  \mathds{1}(\phi_{k+1} \in \bs{\Phi}^{\mathrm{key}}),
\end{equation}
which is only activated when the current phase matches one of the keyframe phases $\bs{\Phi}^{\mathrm{key}}$. This design avoids over-constraining the motion in global space and instead enforces accurate alignment only at sparse but crucial keyframes. To stabilize training under this sparse signal, we follow prior works~\cite{zargarbashi2024robotkeyframing,wang2025beamdojo} and introduce a separate value function $V_\mathrm{track}^{\mathrm{sparse}}$ to better estimate the return from such sparse rewards.
\begin{table}[t]
\centering
\caption{\textbf{Adaptation ranges} of seven tasks during training and testing.}
\setlength{\tabcolsep}{2pt}
\resizebox{1\linewidth}{!}{%
\begin{tabular}{ l | c| c  c c c}
\hline
\multirow{2}{*}{Task} & \multirow{2}{*}{\shortstack{Raw \\ Data}}& \multicolumn{2}{c}{Simulation Train/Test (m)} & \multicolumn{2}{c}{Hardware Test (m)}  \\ \cline{3-6}
& & \textcolor{blue}{Easy} & \textcolor{red}{Hard} & \textcolor{blue}{Easy} & \textcolor{red}{Hard}. \\ \hline
Far Jump $\rightarrow$ & 1.1m & [0.7, 1.6] & [0.2, 0.7] $\cup$  & \{0.7, 1.0\}  & \{1.2, 0.4\} \\
High Jump $\uparrow$ & 0.2m & [0.1, 0.35] & [0.05, 0.1]  $\cup$ [0.35, 0.6] & \{0.1, 0.2\} & \{0.3, 0.4\} \\
Triple Jump $\rightarrow$ & 2.4m & [1.65, 3.15] & [1.2, 1.65] $\cup$ [3.15, 4.2] & \{2.1, 2.7\} & \{1.5, 3.3\}\\
Step Jump $\updownarrow$ & 0.2m & [0.05, 0.25] & [0.25, 0.5] & \{0.2\} & \{0.3\} \\
Tennis $\rightarrow$ & 1.0m & [0.8, 1.2] & [1.2, 1.7] & \{0.8, 1.2\} & \{1.4, 1.6\}\\
Badminton $\searrow$ & 1.3m & [1.3, 1.9] & [1.9, 2.5] & \{1.3, 1.6\} & \{1.9, 2.2\} \\ 
\hline
\end{tabular}
}
\label{tab:adaptaion_range}
\vspace{-0.1in}
\end{table}

\subsubsection{Dense local reward}
In addition, we design a dense local reward to preserve the local joint-space consistency with the reference motion. This reward encourages the reproduced motion to mimic fine-grained patterns from the references:
\begin{equation}
    r_k^{\mathrm{local}} = \mathcal{R}^\mathrm{local}_{\mathrm{track}}(\bs{q}_{\phi_{k+1}}^\mathrm{local},  \hat{\bs{q}}^{\mathrm{local}}_{\phi_{k+1}}),
\end{equation}
A separate value function $V_\mathrm{track}^{\mathrm{local}}$ is employed to estimate the return induced by this dense signal, complementing the sparse global reward. The two value functions $\bs{V}_\mathrm{track}=(V_\mathrm{track}^{\mathrm{sparse}} ,V_\mathrm{track}^{\mathrm{dense}})$ are optimized separately following~\cite{mysore2022multi,huang2025learning}. All reward functions are listed in~\cref{tab:reward}.

\begin{table}[t]
\centering
\caption{\textbf{Reward functions.} The tracking rewards largely follow~\cite{he2025asap} and are categorized into sparse and dense reward groups, with additional regularization rewards to ensure hardware deployability.}
\setlength{\tabcolsep}{6pt}
\resizebox{0.9\linewidth}{!}{%
\begin{tabular}{ l  c  l  c }
\hline
Term & Weight & Term & Weight \\ \hline 
\multicolumn{4}{c}{Sparse Reward $\rightarrow$ $\bs{V}^{\mathrm{sparse}}$}  \\ \hline
Global body position & 10 & Global body rotation & 5 \\
Global feet position & 10 & Termination & -200 \\ \hline 
\multicolumn{4}{c}{Dense Reward $\rightarrow$ $\bs{V}^{\mathrm{dense}}$} \\ \hline
Local body position & 0.75 & Local body rotation & 0.5 \\
Local DoF position & 0.75 & Feet orientation & -5e$^{\text{-2}}$ \\
DoF acceleration & -2.5e$^{\text{-7}}$ & DoF velocity & -5e$^{\text{-4}}$ \\
Action rate & -5e$^{\,\text{-1}}$ & Smoothness & -1e$^{\text{-2}}$ \\
Torques & -1e$^{\,\text{-6}}$ &  Torque limits & 5 \\
DoF position limits & -10 & DoF velocity limits & -5\\ \hline
\end{tabular}
}
\label{tab:reward}
\vspace{-0.1in}
\end{table}

\subsection{Stage 2: Adapters Learning with Adaptive Phase Interval}\label{subsec:training-stage2} 
While stage 1 provides a good policy initialization, its adaptation ability to edited motions remains limited. We posit that the fixed phase interval is a key limitation. As illustrated in \cref{fig:stage2-motivation}, this rigidity results in substantial global and local errors, especially when the required adaptation is large. In practice, such errors may manifest as unnatural pacing or artifacts such as unstable landings. 

\subsubsection{Phase adapter $\phaseadapter$} 
This motivates our design of the phase adapter $\phaseadapter$. It is inspired by the time-warping mechanism in classical motion path editing works, where motions are temporally re-parameterized to preserve naturalness and pacing under spatial edits~\cite{gleicher2001motion,lee2021learning,hsu2007guided}. Formally, the adapter takes the observation as input and outputs a delta phase interval $\Delta\phi^\Delta_k$, which is added to the base interval $\Delta\phi_k$ to obtain an adaptive phase interval $\adaptivephase$:  
\begin{equation}\label{eq:phaseadapter}
    \adaptivephase = \Delta \phi_k + \deltaphase, 
    \quad \deltaphase \sim \phaseadapter(\cdot|\bs{o}_k).
\end{equation}

\begin{table*}[htbp]
    \centering
    \setlength{\tabcolsep}{1.5pt}
    \renewcommand\arraystretch{1.25}
    \caption{\textbf{Main simulation results.} \ours\ is compared against multiple baselines adapted to our problem setting and its ablated versions in a fair setup. Results report mean and standard deviation over three evaluations, each comprising more than ten thousand simulation episodes.}
    \captionsetup{justification=centering, singlelinecheck=false}
    \resizebox{\textwidth}{!}{
        \begin{threeparttable}
            \begin{tabular}{l|ccc|cccc|cccc|cccc}
                \toprule
                \textbf{Comparison Methods} 
                & \multicolumn{3}{c|}{\textbf{Components}}
                & \multicolumn{4}{c|}{\textbf{\textcolor{blue}{Easy Adaptation}}}
                & \multicolumn{4}{c|}{\textbf{\textcolor{red}{Hard Adaptation}}}
                & \multicolumn{4}{c}{\textbf{Overall}}
                \\ 
                \hline
                \ourrow (a) Baselines & $ r^{\mathrm{global}}$ & $\mathcal{D}_{\mathrm{ref}}$ & $\Delta\phi^{\mathrm{ada}}$ 
                & $\metsuccess\uparrow$  & $\metlocal\downarrow$ & $\metglobal\downarrow$ &
                $\metsmooth\downarrow$
                
                & $\metsuccess\uparrow$  & $\metlocal\downarrow$ & $\metglobal\downarrow$ &
                $\metsmooth\downarrow$
                
                & $\metsuccess\uparrow$  & $\metlocal\downarrow$ & $\metglobal\downarrow$ &
                $\metsmooth\downarrow$
    
                \\ \hline
                AMP-Style
                    & Sparse & $\mathcal{D}_{\mathrm{ref}}^{\mathrm{init}}$ & $\thickcdot$
                    & 95.5\%\ci{0.1\%} & 44.5\ci{0.0} & 211.2\ci{0.1} & 19.1\ci{0.3} &
                    70.3\%\ci{0.0\%} & 44.5\ci{0.0} & 247.9\ci{0.1} & 22.3\ci{0.3} & 
                    82.7\%\ci{0.0\%} & 44.5\ci{0.0} & 229.8\ci{0.2} & 20.7\ci{0.3}
                \\
                AMP-Mimic
                    & Sparse & $\mathcal{D}_{\mathrm{ref}}^{\mathrm{init}}$ & $\thickcdot$
                    & 96.8\%\ci{0.0\%} & 35.8\ci{0.0} & 164.4\ci{0.2} & 19.9\ci{0.4} &
                    62.3\%\ci{0.0\%} & 35.7\ci{0.0} & 190.3\ci{0.1} & 21.3\ci{0.4} 
                    & 79.1\%\ci{0.0\%} & 35.7\ci{0.0} & 177.9\ci{0.1} & 20.6\ci{0.4}
                \\
                DeepMimic-NoAdapt
                    & Dense & $\mathcal{D}_{\mathrm{ref}}^{\mathrm{init}}$ & $\thickcdot$
                    & 92.6\%\ci{0.0\%} & 36.6\ci{0.0} & 205.1\ci{0.1} & 17.6\ci{0.6}
                    & \textbf{81.0}\%\ci{0.0\%} & 36.7\ci{0.0} & 484.6\ci{0.2} & 19.4\ci{0.4} 
                    & 86.8\%\ci{0.0\%} & 36.6\ci{0.0} & 351.8\ci{0.2} & 18.6\ci{0.5}
                \\   
                DeepMimic-Adapt
                    & Dense & $\mathcal{D}_{\mathrm{ref}}^{\mathrm{rule}}$ & $\thickcdot$
                    & 95.8\%\ci{0.0\%} & 33.3\ci{0.0} & 123.6\ci{0.0} & \textbf{15.1}\ci{0.3} 
                    & 74.8\%\ci{0.0\%} & 33.3\ci{0.0} & 142.8\ci{0.1} & \textbf{16.7}\ci{0.3} 
                    & 85.1\%\ci{0.0\%} & 33.3\ci{0.0} & 133.5\ci{0.1} & \textbf{15.9}\ci{0.3}
                \\   
                DeepMimic-Adapt-$\Delta\phi^{\mathrm{ada}}$\
                    & Dense & $\mathcal{D}_{\mathrm{ref}}^{\mathrm{rule}}$ & \Checkmark
                    & 93.7\%\ci{0.0\%} & 38.7\ci{0.0} & 170.9\ci{0.0} & 17.2\ci{0.5}
                    & 70.0\%\ci{0.0\%} & 38.7\ci{0.0} & 194.2\ci{0.0} & 17.8\ci{0.8} 
                    & 81.4\%\ci{0.1\%} & 38.8\ci{0.0} & 182.8\ci{0.0} & 17.4\ci{0.5}
                \\   
                \ours-Dense
                    & Dense & $\mathcal{D}_{\mathrm{ref}}^{\mathrm{rule}}$ & \Checkmark
                    & 98.4\%\ci{0.0\%}& 36.3\ci{0.0} & 107.6\ci{0.0} & 17.6\ci{0.1}
                    & 78.0\%\ci{0.0\%} & 36.2\ci{0.0} & 124.7\ci{0.0} & 20.0\ci{0.1} 
                    & \textbf{88.0}\%\ci{0.0\%} & 36.2\ci{0.0} & 115.0\ci{0.0} & 18.8\ci{0.1}
                \\   
                 \textbf{\ours (ours)}
                    & Sparse & $\mathcal{D}_{\mathrm{ref}}^{\mathrm{edit}}$ & \Checkmark 
                    & \textbf{99.6\%}\ci{0.0\%} & \textbf{30.3}\ci{0.0} & \textbf{87.9}\ci{0.0} & 16.0\ci{0.2}
                    & 74.2\%\ci{0.0\%} & \textbf{30.3}\ci{0.0} & \textbf{99.8}\ci{0.1} & 17.3\ci{0.2} 
                    & 86.8\%\ci{0.0\%} & \textbf{30.3}\ci{0.0} & \textbf{94.8}\ci{0.0} & 16.6\ci{0.2}
                \\ \hline
                \ourrow (b) Ablations & Stage 2 & Freeze  & $\Delta\phi^{\mathrm{ada}}$ 
                & $\metsuccess\uparrow$  & $\metlocal\downarrow$ & $\metglobal\downarrow$ &
                $\metsmooth\downarrow$
                
                & $\metsuccess\uparrow$  & $\metlocal\downarrow$ & $\metglobal\downarrow$ &
                $\metsmooth\downarrow$
                
                & $\metsuccess\uparrow$  & $\metlocal\downarrow$ & $\metglobal\downarrow$ &
                $\metsmooth\downarrow$
                
                \\ \hline
                \ours-Stage1
                    & $\thickcdot$ & $\thickcdot$ & $\thickcdot$
                    & 96.7\%\ci{0.0\%} & 43.4\ci{0.0} & 188.1\ci{0.0} & 17.8\ci{0.4} 
                    & 75.2\%\ci{0.0\%} & 43.4\ci{0.0} & 211.8\ci{0.1} & 18.7\ci{0.4} 
                    & 85.7\%\ci{0.0\%} & 43.4\ci{0.0} & 200.4\ci{0.1} & 18.2\ci{0.4}
                \\
                \ours-Stage1-$\Delta\phi^{\mathrm{ada}}$
                    & $\thickcdot$ & $\thickcdot$ & \Checkmark
                    & 92.7\%\ci{0.0\%} & 45.3\ci{0.0} & 195.0\ci{0.1} & 19.8\ci{0.3} 
                    & \textbf{83.2\%}\ci{0.0\%} & 45.3\ci{0.0} & 219.8\ci{0.0} & 20.5\ci{0.2} 
                    & \textbf{88.0\%}\ci{0.0\%} & 45.3\ci{0.0} & 208.1\ci{0.0} & 20.2\ci{0.2}
                \\
                \ours-NoFreeze
                    & \Checkmark & $\thickcdot$ & \Checkmark
                    & 85.0\%\ci{0.0\%} & 44.8\ci{0.0} & 203.4\ci{0.0} & 25.8\ci{0.1}
                    & 71.2\%\ci{0.0\%} & 43.6\ci{0.0} & 224.2\ci{0.0} & 26.9\ci{0.1}
                    & 77.8\%\ci{0.0\%} & 44.1\ci{0.1} & 214.1\ci{0.1} & 26.4\ci{0.1}
                \\
                 \textbf{\ours (default)}
                    & \Checkmark & \Checkmark & \Checkmark 
                    & \textbf{99.6\%}\ci{0.0\%} & \textbf{30.3}\ci{0.0} & \textbf{87.9}\ci{0.0} & \textbf{16.0}\ci{0.2}
                    & 74.2\%\ci{0.0\%} & \textbf{30.3}\ci{0.0} & \textbf{99.8}\ci{0.1} & 17.3\ci{0.2} 
                    & 86.8\%\ci{0.0\%} & \textbf{30.3}\ci{0.0} & \textbf{94.8}\ci{0.0} & \textbf{16.6}\ci{0.2}
                \\
                \bottomrule
            \end{tabular}
        \end{threeparttable}
    }

    \label{tab:simulation_result}
    \vspace{-0.15in}
\end{table*}

\subsubsection{Tracking adapter $\trackingadapter$} 
Given the adaptive phase interval, we need a tracking adapter $\trackingadapter$ to compensate for the tracking action to track the next-step reference motion:
\begin{equation}\label{eq:trackingadapter}
    \adaptivetrack = \bs{a}_k + \deltaphase\cdot\deltatrack, 
    \quad \deltatrack \sim \trackingadapter(\cdot|\bs{o}_k,\adaptivephase),
\end{equation}
where the scaling by $\deltaphase$ ensures that when the delta phase interval is zero, the adaptive action degenerates to the original tracking action. This design eases optimization empirically.

\begin{figure}[t]
    \centering
    \includegraphics[width=1\linewidth]{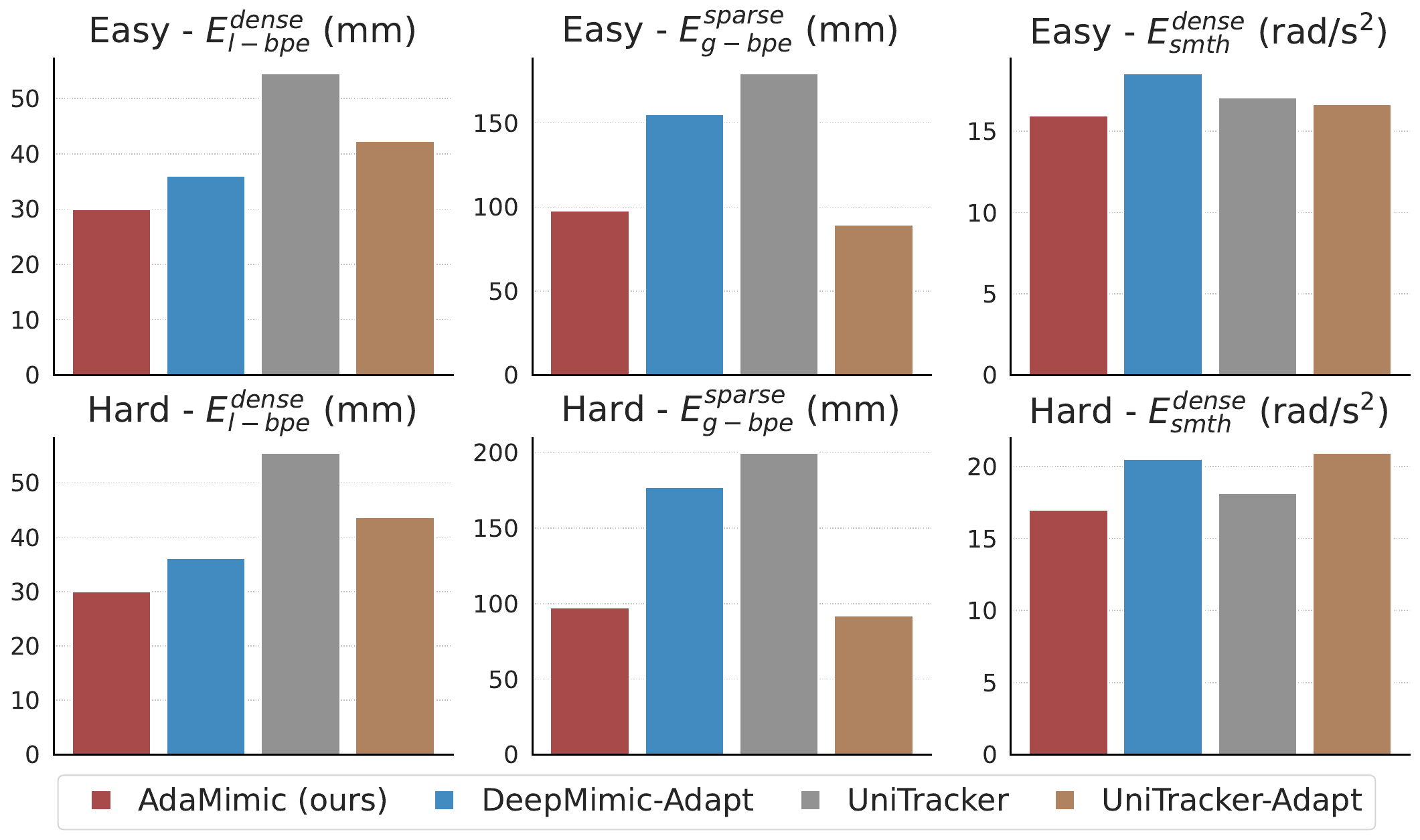}
    \caption{\textbf{Baselines trained with large-scale motion data.} Across five tasks, \ours\ achieves better performance than UniTracker~\cite{Yin2025UniTrackerLU} and its variant adapted with the rule-based motions from DeepMimic-Adapt. Besides, the results indicate that additional motion data does not provide UniTracker with obvious gains in specialization over DeepMimic-Adapt.}
    \label{fig:unitracker}
    \vspace{-0.1in}
\end{figure}
\begin{figure}[t]
    \centering
    \includegraphics[width=1\linewidth]{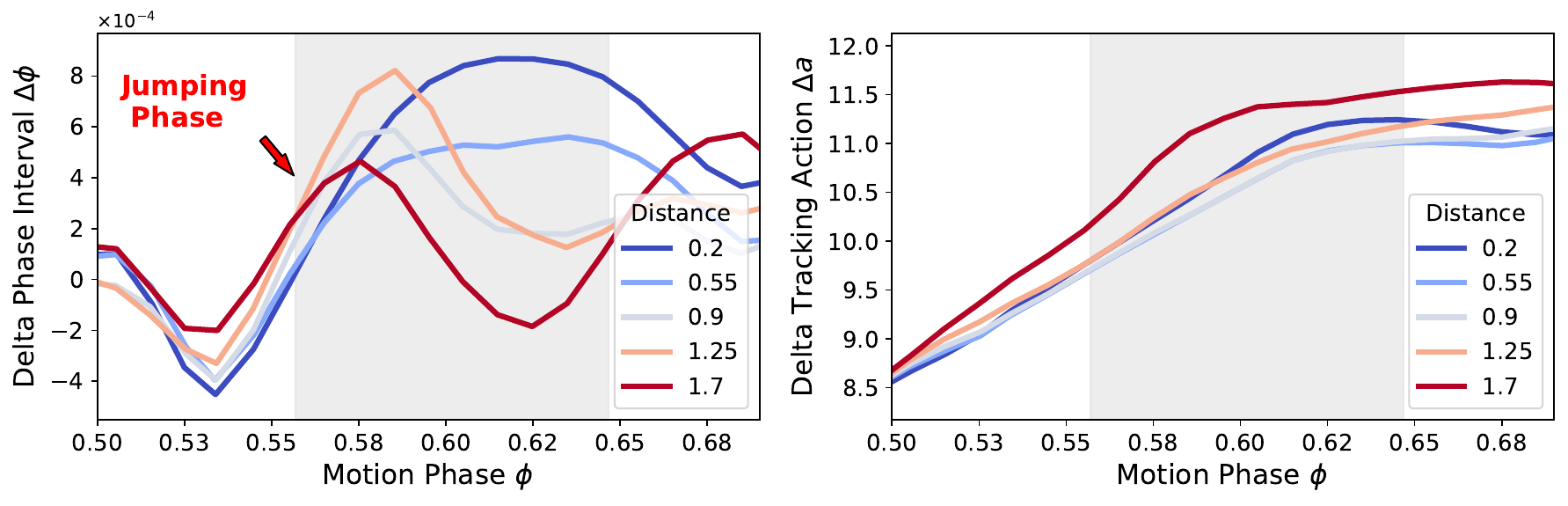}
    \vspace{-0.05in}
    \includegraphics[width=1\linewidth]{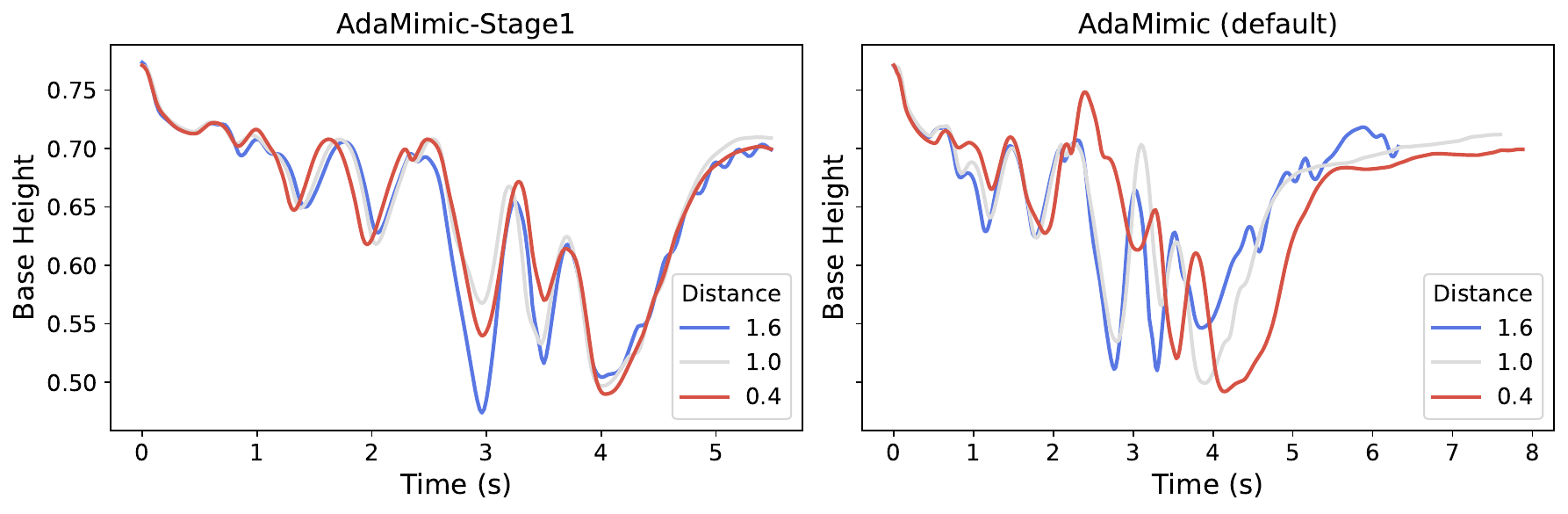}
    \caption{\textbf{Effectiveness of adapters.} (Top) The phase interval and tracking action adapt to different far-jump distances, where longer distances correspond to extended airtime and larger action compensation, while shorter distances lead to reduced adjustments. (Bottom) With adapters, the policy performs effective time warping, adjusting motion speed to improve adaptation.}
    \label{fig:effect_of_adapters}
    \vspace{-0.1in}
\end{figure}

\subsubsection{Optimization}
During training, the two adapters $\bs{\pi}^\Delta=(\phaseadapter,\trackingadapter)$ are paired with separate double critics $\bs{V}^\Delta=(\bs{V}^\Delta_\mathrm{phase}, \bs{V}^\Delta_\mathrm{track})$ to estimate the sparse and dense rewards described in \cref{subsec:training-stage1}. The overall objective is
\begin{equation}\label{eq:objective-stage2}
    \max_{\textcolor{blue}{\bs{\pi}^\Delta}} \;\; 
    \mathbb{E}_{\textcolor{blue}{\bs{\pi}^\Delta}}\left[ \sum_{k=0} \gamma^k \, (\bs{w}\cdot \bs{r}_k) \;\middle|\; \Delta\phi_k,\hat{\bs{q}} \sim \textcolor{blue}{\mathcal{D}_{\mathrm{ref}}^{\mathrm{edit}}} \right].
\end{equation}
Finally, the resulting adaptive action $\adaptivetrack$ is then applied to control the robot to perform adaptive motion tracking.

\subsection{Real-World Deployment}\label{subsec:real-deployment}
We directly deploy the trained policies on the Unitree 29-DoFs G1 humanoid robot. For hardware stability, the waist pitch and roll joints are locked. To enhance robustness, the observation is augmented with a 5-step history. Global localization is realized by lidar odometry through FastLIO~\cite{xu2021fast}. The policy, low-level control, and odometry modules operate at 50Hz, 500Hz, and 10Hz, respectively.

\subsection{Implementation Details}\label{subsec:implementation-details}
The human videos are translated into SMPL motions using GVHMR~\cite{Shen2024WorldGroundedHM} for retargeting. Training is conducted in the Isaac Gym simulator~\cite{makoviychuk2021isaac} with 4096 parallel environments, employing PPO~\cite{schulman2017proximal} as the RL algorithm. Both the policy and value functions are parameterized as 3-layer MLPs. Each episode is initialized from a randomly sampled keyframe~\cite {peng2018deepmimic}. To improve stability for real-world deployment, we adopt L2C2 regularization~\cite{Kobayashi2022L2C2LL,huang2025learning}. The reward weight vector $\bs{w}$ is set to $(1, 0.5)$ for the sparse and dense reward groups, respectively. The adaptive phase interval is defined as $\deltaphase \in [-0.75\Delta \phi_k, \phi_k]$. Finally, PD controllers are configured following~\cite{huang2025learning} for improved simulation performance, while employing configurations from~\cite{Liao2025BeyondMimicFM} during hardware deployment to improve safety and smoothness.

\begin{table*}[htbp]
    \centering
    \setlength{\tabcolsep}{1.5pt}
    \renewcommand\arraystretch{1.25}
     \caption{\textbf{Main hardware results.} In the absence of accurate odometry, we report success rate, local joint tracking error, and smoothness to quantitatively compare \ours\ with three representative baselines. The results indicate that \ours\ achieves strong hardware deployability, particularly in challenging adaptation cases, benefiting from the proposed tracking objectives, motion editing, and adapters. '/' indicates a complete failure in that case.}
    \captionsetup{justification=centering, singlelinecheck=false}
    \resizebox{\textwidth}{!}{
    \begin{threeparttable}
  \begin{tabular}{l|ccc |ccc| ccc | ccc | ccc}
    \toprule
    \multirow{2}{*}{\textcolor{blue}{\textbf{Easy Adaptation}}} &
      \multicolumn{3}{c|}{\textbf{Far Jump}} &
      \multicolumn{3}{c|}{\textbf{High Jump}} &
      \multicolumn{3}{c|}{\textbf{Triple Jump}} &
      \multicolumn{3}{c|}{\textbf{Tennis Hit}} &
       \multicolumn{3}{c}{\textbf{Badminton Hit}} \\  [-0.5ex]

     & $\mathrm{Succ.}\!\uparrow$ & $E^\mathrm{dense}_{\mathrm{l-dof}}\!\downarrow$ & $E^\mathrm{dense}_{\mathrm{smth}}\!\downarrow$ 
     & $\mathrm{Succ.}\!\uparrow$ & $E^\mathrm{dense}_{\mathrm{l-dof}}\!\downarrow$ & $E^\mathrm{dense}_{\mathrm{smth}}\!\downarrow$
     & $\mathrm{Succ.}\!\uparrow$ & $E^\mathrm{dense}_{\mathrm{l-dof}}\!\downarrow$ & $E^\mathrm{dense}_{\mathrm{smth}}\!\downarrow$
     & $\mathrm{Succ.}\!\uparrow$ & $E^\mathrm{dense}_{\mathrm{l-dof}}\!\downarrow$ & $E^\mathrm{dense}_{\mathrm{smth}}\!\downarrow$
     & $\mathrm{Succ.}\!\uparrow$ & $E^\mathrm{dense}_{\mathrm{l-dof}}\!\downarrow$ & $E^\mathrm{dense}_{\mathrm{smth}}\!\downarrow$\\
    \hline
    AMP-Style   
        & 4$/$6 & 35.2\ci{0.2} & 42.8\ci{2.2} 
        & 5$/$6 & 34.8\ci{0.7} & 37.4\ci{7.9} 
        & 4$/$6 & 32.6\ci{0.9} & 41.7\ci{3.5} 
        & 5$/$6 & 31.7\ci{0.5} & 52.9\ci{4.5} 
        & 0$/$6 & / & / \\
    DeepMimic-Adapt 
        & 4$/$6 & 34.4\ci{0.1} & 41.5\ci{1.0} 
        & 1$/$6 & 32.8\ci{0.0} & 65.6\ci{0.0}  
        & \textbf{6$/$6} & 34.0\ci{0.9} & 49.0\ci{3.3} 
        & \textbf{6$/$6} & 32.1\ci{0.1} & 36.5\ci{1.0} 
        & \textbf{6$/$6} & 30.4\ci{0.2} & 50.2\ci{2.2} \\
    \ours-Stage1 
        & \textbf{6$/$6} &  \textbf{33.8}\ci{0.2} & 43.7\ci{2.8} 
        & \textbf{6$/$6} & 32.7\ci{0.1} & 35.7\ci{1.5} 
        & \textbf{6$/$6} & 32.6\ci{0.6} & 54.4\ci{6.3} 
        & \textbf{6$/$6}  & 31.6\ci{0.1} & 33.9\ci{0.9}
        & 5$/$6 & \textbf{28.6}\ci{0.2} & 44.5\ci{3.5}\\  
    \ourrow\textbf{\ours (our)} 
        & 5$/$6 & 35.2\ci{0.7} & \textbf{38.7}\ci{1.4} 
        & \textbf{6$/$6} & \textbf{30.5}\ci{0.3} & \textbf{28.7}\ci{3.4}
        & \textbf{6$/$6} & \textbf{30.7}\ci{0.3} & \textbf{31.3}\ci{6.7}  
        & \textbf{6$/$6} & \textbf{30.7}\ci{0.2} 
        & \textbf{31.9}\ci{1.8} 
        & \textbf{6$/$6} & 28.7\ci{0.1} & \textbf{36.4}\ci{2.0} \\
    \hline
    \multirow{2}{*}{\textcolor{red}{\textbf{Hard Adaptation}}} &
      \multicolumn{3}{c|}{\textbf{Far Jump}} &
      \multicolumn{3}{c|}{\textbf{High Jump}} &
      \multicolumn{3}{c|}{\textbf{Triple Jump}} &
      \multicolumn{3}{c|}{\textbf{Tennis Hit}} &
       \multicolumn{3}{c}{\textbf{Badminton Hit}} \\  [-0.5ex]
     & $\mathrm{Succ.}\!\uparrow$ & $E^\mathrm{dense}_{\mathrm{l-dof}}\!\downarrow$ & $E^\mathrm{dense}_{\mathrm{smth}}\!\downarrow$ 
     & $\mathrm{Succ.}\!\uparrow$ & $E^\mathrm{dense}_{\mathrm{l-dof}}\!\downarrow$ & $E^\mathrm{dense}_{\mathrm{smth}}\!\downarrow$
     & $\mathrm{Succ.}\!\uparrow$ & $E^\mathrm{dense}_{\mathrm{l-dof}}\!\downarrow$ & $E^\mathrm{dense}_{\mathrm{smth}}\!\downarrow$
     & $\mathrm{Succ.}\!\uparrow$ & $E^\mathrm{dense}_{\mathrm{l-dof}}\!\downarrow$ & $E^\mathrm{dense}_{\mathrm{smth}}\!\downarrow$
     & $\mathrm{Succ.}\!\uparrow$ & $E^\mathrm{dense}_{\mathrm{l-dof}}\!\downarrow$ & $E^\mathrm{dense}_{\mathrm{smth}}\!\downarrow$\\
    \hline
    AMP-Style   
        & 3$/$6 & 35.0\ci{0.0} & \textbf{31.8}\ci{1.1} 
        & 0$/$6 & / & / 
        & 3$/$6 & 33.7\ci{0.2} & 42.7\ci{8.9}
        & 4$/$6 & 31.6\ci{0.5} & 66.4\ci{7.3} 
        & 0$/$6 & / & / \\
    DeepMimic-Adapt 
        & 3$/$6 & 34.3\ci{0.1} & 46.6\ci{4.6}
        & 0$/$6 & / & / 
        & \textbf{6$/$6} & 33.8\ci{0.1} & 47.7\ci{2.1}
        & 5$/$6 & 31.1\ci{0.1} & 49.5\ci{4.3} 
        & \textbf{6$/$6} & 30.8\ci{0.1} & 55.3\ci{1.7} \\
    \ours-Stage1 
        & 2$/$6 & \textbf{34.1}\ci{0.3} & 34.3\ci{5.4} 
        & 3$/$6 & 32.5\ci{0.1} & 37.1\ci{3.9} 
        & \textbf{6$/$6} & 32.3\ci{0.1} & 38.3\ci{2.3} 
        & \textbf{6$/$6}  & 31.9\ci{0.1} & \textbf{40.9}\ci{2.2} 
        & 5$/$6 & 29.1\ci{0.1} & 50.9\ci{2.7}\\  
    \ourrow\textbf{\ours (our)} 
        &\textbf{5$/$6} & 35.3\ci{0.5} & 46.2\ci{7.7} 
        & \textbf{5$/$6} & \textbf{31.3}\ci{0.5} & \textbf{28.6}\ci{4.2} 
        & \textbf{6$/$6} & \textbf{31.5}\ci{1.9} & \textbf{31.7}\ci{4.6} 
        & \textbf{6$/$6} & \textbf{30.8}\ci{0.1} & 42.9\ci{1.7} 
        & \textbf{6$/$6} & \textbf{28.9}\ci{0.1} & \textbf{44.5}\ci{2.0} \\

    \bottomrule
  \end{tabular}
  \end{threeparttable}
  }
  \label{tab:hardware_result}
  \vspace{-0.1in}
\end{table*}

\begin{figure*}[t]
    \centering
    \includegraphics[width=1\linewidth]{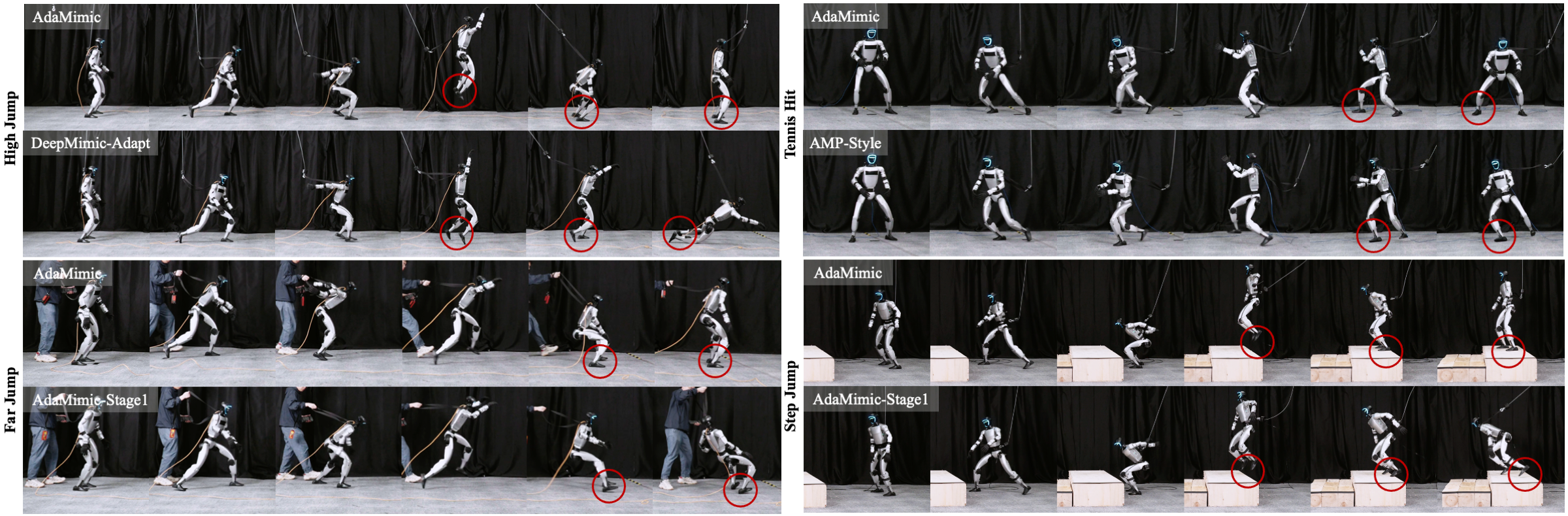}
    \caption{\textbf{Snapshots of real robot motions.} Four representative groups of hardware demonstrations highlight the differences between \ours\ and baseline methods: (1) DeepMimic-Adapt is constrained by physically implausible edited motions used for dense per-frame tracking, (2) AMP-Style exhibits jerky motions and limited imitation accuracy, and (3) \ours-Stage1 yields hardware-unstable policies without the proposed adapters.}
    \label{fig:real_snapshot}
    \vspace{-0.1in}
\end{figure*}

\section{Experiments}
\subsection{Experimental Setup}
\subsubsection{Tasks} We record seven human videos as the evaluation task set, including five various jumping skills and two ball-hitting skills. These tasks are considered agile and difficult enough. Some representative tasks are visualized in~\cref{fig:difficulty-generalization} and their adaptation ranges are presented in~\cref{tab:adaptaion_range}.

\subsubsection{Comparison methods}
We adapt the following baselines to our problem formulation:
\begin{itemize}[leftmargin=4mm]
\vspace{-0.02in}
\item \textbf{Motion as priors:} Adversarial motion priors (AMP-Style;~\cite{peng2021amp}), which use motions as a style regularizer combined with sparse keyframe tracking as the task reward. Its variant, AMP-Mimic, additionally conditions on phase information to better mimic the reference motion.
\item \textbf{Motion tracking from target data:} DeepMimic~\cite{peng2018deepmimic}, including (i) DeepMimic-NoAdapt, which tracks the original motion $\mathcal{D}_{\mathrm{ref}}^{\mathrm{init}}$, and (ii) DeepMimic-Adapt(-$\Delta\phi^{\mathrm{ada}}$), which tracks the reference motions $\mathcal{D}_{\mathrm{ref}}^{\mathrm{rule}}$ generated with a classical linear motion editing method~\cite{gleicher2001motion}.
\item \textbf{Motion tracking from prior motions:}  UniTracker~\cite{Yin2025UniTrackerLU}, trained on large-scale motion datasets, and its adapted variant (UniTracker-Adapt), fine-tuned on $\mathcal{D}_{\mathrm{ref}}^{\mathrm{rule}}$.
\end{itemize}
For fair comparison, we use the same reward functions and observation spaces across all methods, except for UniTracker, which cannot be directly transferred due to its special observations. For this baseline, we use its official implementation.

\subsubsection{Evaluation metrics}
We evaluate each method using four metrics. (1) \textbf{Success rate} considers an episode failed if the average body distance error exceeds 1m at any keyframe phase; unlike prior work, we emphasize keyframes and adopt a looser criterion under the sparse tracking setup. (2) \textbf{Tracking error} measures body distance error in two forms: the \emph{sparse global error} ($\metglobal$, mm), averaged in the world frame at keyframe phases to assess global mimicking accuracy, and the \emph{dense local error} ($\metlocal$, mm), averaged in the root frame over all timesteps to assess local accuracy. (3) \textbf{Smoothness} ($\metsmooth$, rad/s$^2$) is defined as the average joint acceleration over the entire episode. Results are reported as averages over three evaluation runs, each covering more than ten thousand episodes in IsaacGym.

\subsection{Main Simulation Results}
\ours exhibit both great adaptation and tracking precision across all adaptation cases, as shown in~\cref{tab:simulation_result}. 
\paragraphbegin{Comparison with AMP.} The AMP-Style baseline demonstrates moderate adaptability but limited imitation accuracy, primarily due to the absence of phase information in its original formulation. Incorporating phase information alleviates this issue and yields improvements in both adaptability and precision. Nonetheless, the resulting motions remain jerky because of weak motion constraints, and overall performance still falls short of other adaptable methods and  \ours.

\paragraphbegin{Comparison with DeepMimic.} As expected, DeepMimic-noAdapt fails to adapt due to relying on a single reference motion. Its adaptable variants, DeepMimic-Adapt(-$\Delta\phi^{\mathrm{ada}}$), achieve notable performance gains, highlighting the effectiveness and generality of our formulation. However, their tracking performance degrades substantially from easy to hard adaptation scenarios (see also example in~\cref{fig:real_snapshot}), reflecting limitations imposed by the underlying motion path editing. In contrast, our method avoids such restrictive editing assumptions and instead leverages RL with keyframing to produce more physically plausible motions.

\paragraphbegin{Comparison with UniTracker.} As shown in~\cref{fig:unitracker}, we observe that: (1) the pre-trained model, despite being trained on massive motion data, struggles to perform agile motions; (2) finetuning with $\mathcal{D}_{\mathrm{ref}}^{\mathrm{rule}}$ yields some gains but does not surpass DeepMimic-Adapt trained from scratch; and (3) both versions overall underperform \ours. These findings further suggest that universal trackers could be made less dependent on pre-defined data during deployment by incorporating a more physically plausible motion editing module.

\subsection{Analyses of Key Designs}

\paragraphbegin{Effectiveness of adapters.} \cref{tab:simulation_result} and~\cref{fig:stage2-motivation} quantify the benefits of adapters, while~\cref{fig:effect_of_adapters} reveals their mechanism. The top panel shows that phase intervals and delta action scale with jump distance: longer jumps lead to extended airtime and larger compensation, whereas shorter jumps require smaller adjustments, mostly concentrated around landing. Together, this decoupling of timing and correction enables motion speed adaptation without altering the motion pattern, resulting in lower tracking error, smoother landings, and higher success rates, especially in hard adaptation cases.

\paragraphbegin{Keyframe editing outperforms per-frame editing.}  We observe that keyframe-based \ours\ consistently outperforms its per-frame counterpart, \ours-Dense, indicating the advantage of sparse keyframe selection for adaptation. Interestingly, the trend is reversed when comparing \ours-Stage1 with DeepMimic-Adapt, suggesting that the combination of keyframing and adapters is crucial for achieving both precise tracking and great adaptation.

\paragraphbegin{Two-stage training is necessary.} Our motivation for the two-stage design arises from the observation that training with adaptive phase intervals (\ours-Stage1-$\Delta\phi^{\mathrm{ada}}$) but inference with fixed intervals can even degrade performance compared to training with fixed intervals (\ours-Stage1). We hypothesize that this is due to increased optimization difficulty incurred by the varied phase interval, which can be substantially mitigated by our two-stage design.

\paragraphbegin{Freezing the tracking policy is important.} Finetuning the tracking policy during the second stage (\ours-NoFreeze) noticeably degrades both smoothness and tracking accuracy. We attribute this degradation to increased optimization difficulty when updating the policy alongside adapters.

\subsection{Main Hardware Results}
We present snapshots of real robot motions to compare \ours with representative baselines in~\cref{fig:real_snapshot}. DeepMimic-Adapt fails in extreme cases, such as high jumps, because the rule-based augmented motions are physically inconsistent and cannot be reliably executed on hardware. AMP-Style shows noticeable instability in dynamic tasks like tennis hitting: motions are jerky, and forceful strikes lead to poor balance and uncoordinated execution. AdaMimic-Stage1, lacking the proposed adapters, exhibits large sim-to-real gaps; in particular, the ankle joints become highly unstable during forceful motions, reflecting inadequate temporal and action adaptation. 

In contrast, \ours\ combines sparse keyframes with phase and tracking adapters, enabling the policy to adjust motion timing and per-step actions while maintaining original motion patterns. This results in physically plausible and robust execution across all tasks. Quantitatively, as shown in~\cref{tab:hardware_result}, \ours\ consistently achieves high success rates, reduced local joint errors, and smoother motions in both easy and hard adaptation scenarios, highlighting the effectiveness and generality of our method on the real robot.

\section{Conclusion}
We have presented \ours, a novel motion tracking framework for adaptable humanoid control from a single reference motion. Our framework addresses the limitations of prior methods, which either sacrifice tracking accuracy for adaptation or require large-scale reference motions for each adaptation condition. By leveraging augmented sparse keyframes and a two-stage training strategy with phase and tracking adapters, \ours enables accurate imitation while extending adaptability to diverse tasks and conditions. Experimental results on the Unitree G1 humanoid robot demonstrate that our controllers achieve adaptable motions across many tasks, validating both the effectiveness and generality of our framework. Looking forward, \ours holds promise for extending beyond the demonstrated scenarios to more complex and interactive whole-body skills, such as perceptive and professional ball games.

\section{Limitations and Future Directions}
We acknowledge several key limitations that are considered valuable to be addressed in the future.

\paragraphbegin{Wide task scope.} Our method currently focuses on tasks with clear parametric forms (e.g., jump or strike distances), while many skills lack such representations. Extending to less structured tasks could broaden its applicability.

\paragraphbegin{General motion editing mechanism.} Keyframe selection and editing are manually specified, which restricts scalability. Developing automatic editing mechanisms could make the framework more general and efficient.

\paragraphbegin{Utilization of massive motion data.} Universal trackers trained on large datasets still underperform, even with finetuning. Better strategies to exploit massive motion data are valuable for stronger adaptation.

\paragraphbegin{Adaptation for interactive tasks.} Our framework executes motions without environment feedback, limiting interactivity. Incorporating perception will enable adaptive and responsive behaviors in interactive tasks such as ball games.

{\footnotesize
\bibliographystyle{IEEEtran}
\bibliography{references.bib}
}

\appendix\label{appendix}

\subsection{Task Details}
\paragraphbegin{Motion editing.} For each task, we selected two types of keyframes: (1) Keyframes with semantics. For example, the frame before and after jumping. (2) Keyframes without semantics. These keyframes are uniformly sampled from a clip of motion to ensure smoothness and tracking accuracy. The global poses of the former keyframes are further edited in one dimension. For motion details, please refer to the   \href{https://github.com/InternRobotics/AdaMimic}{code}.

\paragraphbegin{Observations} are composed of the following parts:
\begin{itemize}[leftmargin=4mm]
\vspace{-0.02in}
\item Base angular velocity, 3 dimensions.
\item Projected gravity $\dot{\bs{\omega}}_k$, 3 dimension.
\item Joint positions $\bs{\omega}_k$, 27 dimension.
\item Joint velocities $\dot{\bs{\theta}}$, 27 dimension.
\item Previous actions $\bs{a}_{k-1}$, 27 dimension.
\item Current phase $\phi$, 1 dimension.
\item Task variable $\psi$, 1 dimension.
\item Lidar odometry, 3 dimensions. The relative position based on the initial lidar coordinate.
\item Base linear velocity, 3 dimension, critic-only.
\item Reference joint positions, 27 dimensions, critic-only.
\end{itemize}

\subsection{Baseline implementations.}  
\begin{itemize}[leftmargin=4mm]
\vspace{-0.02in}
\item \textit{AMP-style} and \textit{AMP-Mimic} use 5-step DoF positions as the discriminator observation. The weight of the style reward is 0.1, while the weight of the task reward is 0.9. The style reward is classified into the sparse reward group. 
\item \textit{DeepMimic}-based baselines use the same rewards as ours. Differently, \textit{DeepMimic-NoAdapt} uses the original motion and therefore has no adaptation ability. \textit{DeepMimic-Adapt} uses the rule-based augmented motions with a fixed phase interval. \textit{DeepMimic-Adapt-$\Delta\phi^{\mathrm{ada}}$} additionally trains a flexible phase interval.
\item \textit{UniTracker} follows its official implementation without modification. It uses the rule-based augmented motions during inference. \textit{UniTracker-Adapt} additionally introduces a fine-tuning stage on the augmented motions. Since training UniTracker in terrains is not straightforward, we only test UniTracker in five tasks that do not depend on terrains. 
\end{itemize}
The method of rule-based motion augmentation mainly follow~\cite{gleicher2001motion}. Specifically, we first determine the keyframes in the reference motion, which are the same as those in the motion processing stage. Then, we linearly adjust the global position of in-between frames without adjusting their local pose. Each task variable $\psi$ results in a corresponding motion, which is collected into $\mathcal{D}_{\mathrm{ref}}^{\mathrm{rule}}$.

\subsection{Hardware Setup}
We use the Unitree 29-DoF G1 humanoid robot (6 per leg, 7 per arm, and 1 in the waist). Waist roll and pitch joints are locked for stability and safety. We used the MID360 lidar on board with FastLIO~\cite{xu2021fast} to obtain odometry. The PD controller follows the designs in BeyondMimic~\cite{Liao2025BeyondMimicFM}, which significantly improves the hardware deployment in terms of stability, smoothness, and safety. Due to the drift issue of the odometry and the randomly initialized robot's pose, we observe randomness in policy performance. Therefore, we conduct multiple tests of each policy in each task for statistical significance.

\subsection{Training Details}
\paragraphbegin{Domain randomization.} We apply domain randomization during training in the simulation. The randomization terms largely follow HoST~\cite{huang2025learning}, which are sufficient for overcoming the sim-to-real gap.~\cref{table:domain_randomization} below lists each term:
\begin{table}[h]
    \centering
    \vspace{-0.05in}
    \setlength{\tabcolsep}{10pt}
    \caption{Domain randomization for adaptive motion tracking.}
    \begin{tabular}{ll}
    \toprule 
    Term & Value \\
    \midrule 
    Trunk Mass & $\mathcal{U}(-2,5)$kg  \\
    Base CoM offset & $\mathcal{U}(-0.1,0.1)$m (XYZ)\\
    Link mass & $\mathcal{U}(0.9,1.1)\times$ default kg  \\ 
    Fiction & $\mathcal{U}(0.1,1.1)$ \\
    Restitution & $\mathcal{U}(0,0.1)$ \\
    P Gain & $\mathcal{U}(0.85,1.15)$ \\
    D Gain & $\mathcal{U}(0.85,1.15)$ \\
    Motor Strength & $\mathcal{U}(0.9,1.1)$ \\
    Control delay & $\mathcal{U}(0, 100)$ms \\
    \bottomrule
    \end{tabular}
    \label{table:domain_randomization}
\end{table}

\paragraphbegin{Termination conditions.} We follow the termination conditions in ASAP~\cite{he2025asap}, including termination curriculum and termination terms. Beyond that, given the nature of keyframe tracking, we observe that very strict termination conditions will impede the exploration of highly dynamic parts of the motion (\eg jumping). We therefore relax the conditions after 6000 training epochs to ensure successful learning of the whole motion.

\paragraphbegin{Reward scales.} We observe that the identical reward scales of each keyframe in the sparse reward group will lead to conservative policies, \eg the policy learns to avoid jumping to collect more local rewards. To overcome this problem, we assign some keyframes with semantics with a higher reward scale to encourage exploration of more dynamic behaviors.

\paragraphbegin{PPO hyperparameters} are listed in~\cref{tab:hyperparameters}:
\begin{table}[!h]
    \centering
    \vspace{-0.05in}
    \caption{Hyperparameters of PPO.}
    \vspace{-0.05in}
    \begin{tabular}{ll}
    \toprule
    \textbf{Hyperparameter} & \textbf{Value} \\

    \midrule
    Number of envs & 4096 \\
    Number of steps per iteration & 75 \\
    Number of learning epochs & 5 \\
    Clip range & 0.2 \\
    Entropy coefficient & 0.01 \\
    GAE balancing factor $\lambda$ & 0.95 \\
    Desired KL-divergence & 0.01 \\
    Actor and double critic MLP & [512, 256, 128] \\
    Discount factor & Sparse 1; Dense 0.99 \\
    Initial learning rate & 1$e^{-3}$\\
    \bottomrule
    \end{tabular}
    \label{tab:hyperparameters}
\end{table}

\end{document}